\documentclass{article}
\usepackage{spconf,amsmath,graphicx,booktabs,cite}
\usepackage{hyperref}

\title{From Weak Task Specifications to Scientific Extraction Agents: Optimizing Task Construction}

\name{
Zixiao Dong$^{1,3}$,
Wei Yang$^{2,3}$,
Zihao Liu$^{1,3}$,
Chenshu Li$^{1,3}$,
Longzhang Liu$^{1,3}$,
Tao Tan$^{4}$,
Hong Xie$^{1,3*}$
\thanks{
$*$ Corresponding author: Hong Xie (hongx87@ustc.edu.cn).
}
}

\address{
$^{1}$School of Computer Science and Technology, University of Science and Technology of China\\
$^{2}$University of Science and Technology of China, 
$^{3}$State Key Laboratory of Cognitive Intelligence\\
$^{4}$CCCC Second Highway Consultants Co., Ltd.
}

\begin{document}
\ninept
\maketitle

\begin{abstract}
Most methods that optimize LLM prompts and agent workflows assume that task-specific output schemas, extraction instructions,
and evaluation criteria are predefined. For scientific extraction agents, however, a short task goal may not fully determine these
components, while specifying them manually is costly. We study the upstream problem of constructing the task-specific configuration
from a weak specification containing only a short goal and unannotated reference documents. Rather than treating automatic
construction as a fixed preprocessing step, our framework constructs a task-specific schema, extraction instructions, and base
training rubrics, then keeps schema construction and extraction instructions editable during optimization. Failure-focused updates
concentrate textual-gradient feedback on lower-scoring documents, while training-time evaluation criteria adapt to recurring failures.
On a heterogeneous-catalysis literature corpus, automatic construction remains improvable, and optimizing both schema construction
and extraction instructions performs best across all four judge--rubric settings, with ablations and blinded human evaluation
supporting the proposed formulation.
\end{abstract}


\section{Introduction}

Scientific extraction agents convert full papers into structured, task-specific records.
This is difficult because relevant evidence can be distributed across body text, tables, captions, and distant sections \cite{gururaja2025collage,yang2026agentcat}, while
extracted information often depends on relations and contextual qualifiers needed for correct interpretation \cite{mavracic2021cde2,jain2020scirex}. 
In practice, such an agent also depends on task-specific choices about the output structure, extraction instructions, and evaluation criteria.
Users may know the information they seek without being able to fully specify these components in advance, while manually constructing them often requires
substantial effort and domain knowledge. We therefore define a \emph{weak task specification} as a short goal paired with unannotated reference documents, without a predefined task configuration.

Prompt optimization methods revise task-specific instructions using model feedback or search \cite{pryzant2023protegi,yang2024opro,yuksekgonul2024textgrad}, while program optimizers extend this idea to modular and multi-stage LLM pipelines \cite{khattab2024dspy,opsahl2024mipro}. More recent work optimizes agent-level functions and prompts \cite{zhang2024agentoptimizer,spiess2025autopdl,agrawal2026gepa}. Existing prompt and program optimizers typically improve components within an already instantiated task-specific system. We instead study the upstream problem of 
constructing that system from a weak task specification alone. Because such a specification does not fully determine the task-specific configuration, we treat automatic construction as an initialization rather than a fixed preprocessing step.

We define \emph{automatic task construction} as generating a task-specific schema, extraction instructions, and base training rubrics from a weak task specification. \emph{Failure-focused optimization} then updates the constructed components using lower-scoring documents and adapts the training rubric to recurring failures. Our contributions are:
\begin{itemize}
    \item We formulate task construction itself as an optimization problem under weak task specifications, with editable schema construction and extraction instructions and adaptive training rubrics.
    \item We introduce failure-focused optimization that concentrates updates on poorly performing documents and adapts training-time evaluation criteria to recurring failures, while decoupling adaptive training feedback from frozen-rubric selection.
    \item Experiments on heterogeneous-catalysis literature show that both schema construction and extraction instructions remain improvable after initialization, and that sequentially optimizing them performs best across all four judge--rubric settings, supported by ablations and blinded human evaluation.
\end{itemize}

\section{Related Work}

Earlier scientific information extraction systems relied on schemas, ontologies, rules, and task-specific pipelines \cite{mavracic2021cde2,jain2020scirex,luan2018scierc}. Recent
LLM-based methods support entity--relation, scientific-contribution, and material-property extraction \cite{dagdelen2024structured,shamsabadi2024scientific,polak2024chatextract}.
Systems using task instructions or annotation guidelines extend IE to user-specified or unseen tasks \cite{jiao2023instruct,sainz2024gollie}.
Full-document extraction remains challenging because task-relevant evidence is often distributed across a paper.
Collage \cite{gururaja2025collage} addresses this through document-aware pipeline design, while AgentCAT \cite{yang2026agentcat} combines progressive schema evolution with an evidence-grounded candidate--resolve--review agent workflow.
In contrast, we make task construction itself an optimization target under weak specifications, optimizing both schema construction and extraction instructions.

Automatic prompt optimization revises instructions or language-model programs using model feedback, search, or other optimization signals. ProTeGi \cite{pryzant2023protegi} uses natural-language gradients, and OPRO \cite{yang2024opro} treats an LLM as an optimizer. DSPy and MIPRO extend optimization to modular and multi-stage programs \cite{khattab2024dspy,opsahl2024mipro}, while TextGrad \cite{yuksekgonul2024textgrad} propagates natural-language feedback to editable components. Recent work further extends optimization to agent-level components: AgentOptimizer treats agent functions as learnable parameters \cite{zhang2024agentoptimizer}, AutoPDL optimizes agent prompts \cite{spiess2025autopdl}, GEPA uses reflective prompt evolution \cite{agrawal2026gepa}, and multi-judge optimization combines feedback from several evaluators \cite{zhao2026gmpo}.
These methods, like instruction-based IE \cite{jiao2023instruct,sainz2024gollie}, assume an instantiated task-specific program. We instead make task construction itself an optimization target.

Evaluation criteria may also need to evolve during optimization.
DynamicRubric \cite{wang2026dynamicrubric} co-evolves an evaluator and policy using rubrics conditioned on current-policy responses.
In our setting, stage-specific base training rubrics are generated from the task goal and extended with criteria for recurring failures, while checkpoints are selected using a frozen final-round rubric on held-out documents.

\section{Method}

\begin{figure*}[t]
\centering
\includegraphics[width=\textwidth]{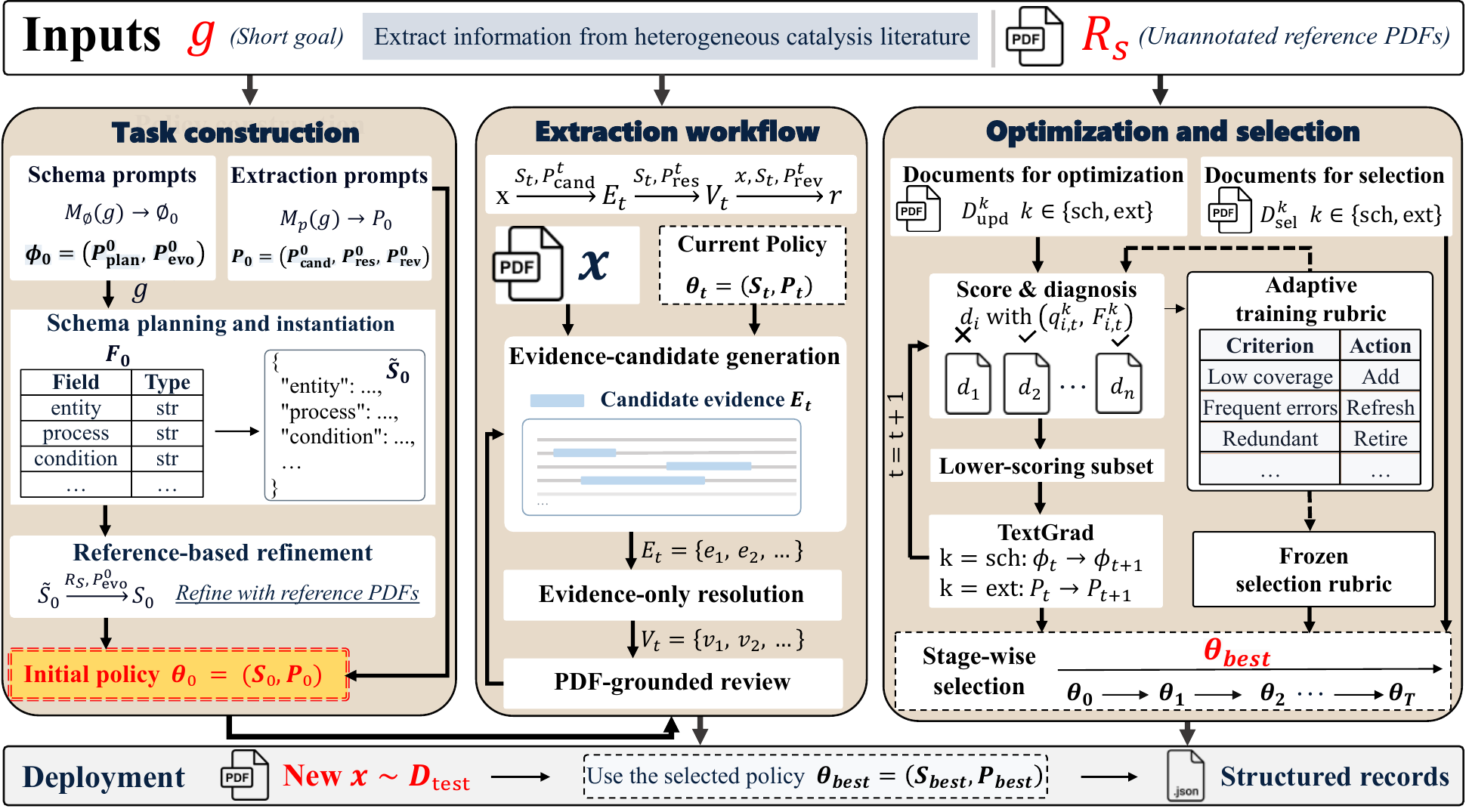}
\caption{Overview of the proposed weak-specification task-construction framework. This framework constructs a task-specific policy from weak specifications, refines it through failure-focused optimization, and selects the final policy using a frozen rubric.}
\label{fig:overview}
\end{figure*}

Given a weak task specification consisting of a short goal $g$ and unannotated reference documents $R_S$, our goal is to construct and optimize a scientific extraction agent by generating an initial runtime schema $S_0$, extraction prompts $P_0$, and stage-specific base training rubrics.
Figure~\ref{fig:overview} summarizes the overall framework.
We denote the deployable extraction policy by $\theta=(S,P)$, where $S$ and $P$ denote the runtime schema and extraction prompts, respectively; the initialized policy is therefore $\theta_0=(S_0,P_0)$.
Optimization operates on two editable components: the schema-construction policy $\phi$ and $P$, with $\phi$ producing $S$. We index optimization rounds by $t\in\{0,\ldots,T\}$, where $t=0$ denotes initialization, and use $k\in\{\mathrm{sch},\mathrm{ext}\}$ for the schema-construction and extraction stages.

\subsection{Schema construction from weak specifications}

At round $t$, schema construction is controlled by
\begin{equation}
\phi_t=(P_{\mathrm{plan}}^t,P_{\mathrm{evo}}^t)
\label{eq:phi}
\end{equation}
Here, $P_{\mathrm{plan}}^t$ and $P_{\mathrm{evo}}^t$ denote the planning and schema-evolution prompts. The fixed, domain-independent meta-prompt $M_\phi$ generates the initial prompts $\phi_0$ and the schema-stage base rubric $J_0^{\mathrm{sch}}$ from $g$. The initial planning prompt maps $g$ to a provisional task framework $F_0$. A deterministic compiler converts $F_0$ into the provisional schema $\widetilde S_0$. The initial evolution prompt then refines this schema against $R_S$:
\begin{equation}
g
\xrightarrow{P_{\mathrm{plan}}^0}
F_0
\xrightarrow{\mathrm{compile}}
\widetilde S_0
\xrightarrow{R_S,\,P_{\mathrm{evo}}^0}
S_0
\label{eq:schemachain}
\end{equation}
At later rounds, $S_t=S(\phi_t;g,R_S)$ denotes the schema regenerated by the current construction prompts.
The goal specifies the task intent, while $R_S$ supplies document-grounded detail for schema refinement.

\subsection{Extraction-prompt generation and execution}

From the short goal $g$, the fixed meta-prompt $M_P$ generates the stage-specific extraction prompts
\begin{equation}
P_0=(P_{\mathrm{cand}}^0,P_{\mathrm{res}}^0,P_{\mathrm{rev}}^0)
\end{equation}
Here, $P_{\mathrm{cand}}^0$, $P_{\mathrm{res}}^0$, and $P_{\mathrm{rev}}^0$ are the candidate-generation, evidence-resolution, and review prompts. The same
initialization step produces the extraction-stage base rubric $J_0^{\mathrm{ext}}$.
These extraction prompts are generated from $g$ independently of $S_0$ and paired with the runtime schema during execution.

During execution, each schema section follows the AgentCAT \cite{yang2026agentcat} workflow of candidate generation, evidence-only resolution, and PDF-grounded review. These stages collect supporting evidence, map it to structured values, and verify and correct the values against the full PDF before assembling the final record.

\subsection{Failure-focused textual-gradient optimization}

We use TextGrad \cite{yuksekgonul2024textgrad} as the underlying prompt optimizer, while modifying the feedback process in two ways: updates focus on lower-scoring documents, and the training-time evaluation criteria can adapt to recurring failures. The set $D_{\mathrm{upd}}^k$ contains the documents used for optimization feedback at stage $k$.
Schema-TG regenerates $S_t$ from the updated construction prompts, whereas Extract-TG jointly updates the editable prompt components in $P_t$ under a fixed schema. For update document $d_i$, the stage evaluator returns score $q_{i,t}^k$ and natural-language diagnosis $F_{i,t}^k$. At the schema stage, it assesses whether the constructed schema can adequately represent the task-relevant information supported by $d_i$, without executing the full extraction workflow. We denote the lower-scoring half of the update set by $\mathcal{H}_t^k$:
\begin{equation}
\mathcal{H}_t^k = \operatorname{BottomHalf}\left(D_{\mathrm{upd}}^k;\{q_{i,t}^k\}\right)
\label{eq:lowerhalf}
\end{equation}
Only diagnoses from $\mathcal{H}_t^k$ are passed to TextGrad, so each update focuses on the currently lower-scoring documents.

Each stage starts from a fixed base rubric $J_0^k$ and maintains an adaptive supplement $\Delta_t^k$:
\begin{equation}
J_t^k=J_0^k\oplus\Delta_t^k
\label{eq:adaptive}
\end{equation}
Here, $\oplus$ augments the base rubric with the active adaptive criteria in $\Delta_t^k$, whose capacity is bounded by
\begin{equation}
|\Delta_t^k|\le C
\label{eq:capacity}
\end{equation}
with $C$ the maximum number of active adaptive criteria. After each round, the evaluator summarizes the observed errors and may update $\Delta_t^k$ with criteria for recurring or consequential failure modes. Redundant criteria refresh existing entries, while inactive criteria may be retired. The adaptive rubric supplies optimization feedback; checkpoint scores are produced separately as described next.

\subsection{Frozen-rubric checkpoint selection}

Since the training rubric evolves across rounds, scores produced under different round-specific rubrics are not directly comparable. This separation prevents checkpoint ranking from being confounded by rubric drift across optimization rounds. After optimization, the final-round training rubric $J_T^k$ is frozen and used as the selection rubric, $J_{\mathrm{sel}}^k := J_T^k$. Every saved checkpoint is then rescored on the held-out set $D_{\mathrm{sel}}^k$ under this common rubric. For a document $d$ and stage artifact $A$, $J_{\mathrm{sel}}^k(d,A)$ returns a scalar score. The function $E(d;S,P_t)$ denotes the record extracted from $d$ with schema $S$ and prompts $P_t$. We save the stage state after each round as checkpoint $t$. Schema and extraction checkpoints are selected separately:
\begin{align}
\phi_{\mathrm{best}}
&= \operatorname*{arg\,max}_{\phi_t}
\frac{1}{|D_{\mathrm{sel}}^{\mathrm{sch}}|}
\sum_{d\in D_{\mathrm{sel}}^{\mathrm{sch}}}
J_{\mathrm{sel}}^{\mathrm{sch}}
\bigl(d,S(\phi_t;g,R_S)\bigr)\\
P_{\mathrm{best}}(S)
&= \operatorname*{arg\,max}_{P_t}
\frac{1}{|D_{\mathrm{sel}}^{\mathrm{ext}}|}
\sum_{d\in D_{\mathrm{sel}}^{\mathrm{ext}}}
J_{\mathrm{sel}}^{\mathrm{ext}}
\bigl(d,E(d;S,P_t)\bigr)
\end{align}
Here, $\phi_{\mathrm{best}}$ is the selected schema-construction checkpoint. The term $P_{\mathrm{best}}(S)$ is the extraction-prompt checkpoint selected for a fixed schema $S$. Let $S_{\mathrm{best}}=S(\phi_{\mathrm{best}};g,R_S)$ and $P_{\mathrm{best}}=P_{\mathrm{best}}(S_{\mathrm{best}})$. The final deployable policy is $\theta_{\mathrm{best}}=(S_{\mathrm{best}},P_{\mathrm{best}})$.

\section{Experiments}

\subsection{Experimental Setup}

The corpus consists of unannotated heterogeneous-catalysis papers spanning experimental catalysis, spectroscopy, density-functional theory, mechanistic analysis, and kinetic modeling. Twelve PDFs are used for schema refinement.
Each optimization stage additionally uses four update documents and two selection documents, with final evaluation on a separate held-out 50-paper target corpus.

GPT-5.6 Terra (temperature 0; high reasoning effort) is used for extraction and optimization-time evaluation, with native PDF inputs and five textual-gradient updates per stage. Auto-init deploys the generated schema and extraction prompts without optimization.
Schema-TG updates schema construction only, and Extract-TG updates extraction instructions only. Joint-TG sequentially optimizes both stages. It first selects an optimized schema-construction checkpoint and then optimizes extraction instructions under the resulting schema.
To isolate stage effects, Extract-TG reuses the Auto-init schema, while Joint-TG reuses the selected Schema-TG schema before extraction-instruction optimization, keeping the schema fixed within each paired comparison. We compare with the original AgentCAT \cite{yang2026agentcat} and an adapted AgentCAT-short variant.
AgentCAT uses a manually designed task specification, extraction prompts, and task-specific evaluation criteria; AgentCAT-short keeps all other AgentCAT components unchanged and replaces only its detailed task specification with our six-word goal.

All methods use the same base model and target corpus. Within each stage, optimized variants share the same five-update budget, document sets, and evaluation protocol, differing only in whether schema construction, extraction instructions, or both are optimized.

The feedback-selection ablation keeps these settings fixed and changes only whether all document diagnoses or the lower-scoring half are passed to the optimizer.
For adaptive training rubrics, we set $C=3$. At each round, the evaluator may propose zero to two new criteria. When the capacity is exceeded, criteria triggered most frequently and recently are retained; a criterion is retired after three inactive rounds.
Code is available at \url{https://github.com/yyhlm/weak-spec-extraction-agents}.

\subsection{Evaluation}

To evaluate the proposed method, we assess full-paper extraction using two expert-designed, fixed multidimensional rubrics.
R1 assesses record-level extraction quality through coverage, accuracy, specificity, and usability, weighted 35\%, 35\%, 20\%, and 10\%, respectively, whereas R2 assesses scientific-context fidelity through causal linking, parameter alignment, mechanistic fidelity, and micro–macro distinction, weighted 30\%, 30\%, 25\%, and 15\%, respectively.
Each dimension is scored on a 0–100 scale, and each rubric score is computed as the weighted average of its dimensions.

Following the LLM-as-a-judge paradigm \cite{zheng2023judge}, two PDF-aware judges evaluate identical extraction files while blinded to method and checkpoint identity.
Judge-T uses GPT-5.6 Terra and Judge-5.5 uses GPT-5.5, both with temperature 0 and high reasoning effort, yielding four judge–rubric evaluation settings.

Blinded human spot-checks are additionally conducted on ten papers spanning five study types. Joint-TG is compared separately with Auto-init and AgentCAT in balanced A/B order, with assessors consulting the source PDFs and independently judging record-level quality and scientific-context fidelity.

\subsection{Main results}

Table~\ref{tab:main} shows the main results. Auto-init averages 78.37, close to AgentCAT-short (79.18) and AgentCAT (79.81), showing that weak-specification task construction is already competitive.
Joint-TG reaches 83.40 and leads all four judge--rubric settings.
Without a manually specified task configuration, Joint-TG surpasses both AgentCAT baselines with manually developed prompts and evaluation rubrics.
Relative to Auto-init, sequential optimization improves the four-score mean by 5.03 points. The stage-wise variants show that both components can be further improved after automatic construction, with extraction-instruction optimization producing the larger single-stage gain.

\begin{table}[ht]
\centering
\caption{Evaluation scores on the complete 50-PDF target corpus.}
\label{tab:main}
\resizebox{\columnwidth}{!}{%
\begin{tabular}{lcccc}
\toprule
Method & T-R1 & T-R2 & 5.5-R1 & 5.5-R2 \\
\midrule
Auto-init & 79.63 & 77.42 & 79.02 & 77.40 \\
AgentCAT-short \cite{yang2026agentcat} & 81.55 & 78.93 & 79.05 & 77.20 \\
AgentCAT \cite{yang2026agentcat} & 81.04 & 80.49 & 79.52 & 78.17 \\
Schema-TG (best) & 81.68 & 81.63 & 80.95 & 79.02 \\
Extract-TG (best) & 83.63 & 82.90 & 81.30 & 80.75 \\
Joint-TG (best/best) & \textbf{84.50} & \textbf{84.38} & \textbf{83.67} & \textbf{81.05} \\
\bottomrule
\end{tabular}}
\end{table}

For Joint-TG versus Auto-init, 95\% CIs are estimated using 100,000 paired percentile bootstrap resamples over the 50 target papers. Under Judge-T, the R1 and R2 improvements are $+4.88$ [1.35, 8.98] and $+6.95$ [3.40, 10.95], respectively. Under Judge-5.5, they are $+4.65$ [1.73, 8.23] and $+3.65$ [0.25, 7.48].
All four 95\% bootstrap intervals lie above zero, indicating consistent gains across both judges and both rubrics.
For Joint-TG versus AgentCAT, the paired 95\% CIs are [1.65, 5.42] and [1.94, 5.91] for Judge-T R1 and R2, respectively. For Judge-5.5, the corresponding intervals are [1.55, 7.23] and [0.63, 6.68].

In both blinded human comparisons, Joint-TG is preferred on all 10 papers for record-level quality and on 9 of 10 for scientific-context fidelity.
For each comparison, two-sided sign tests give $p=0.0020$ and $p=0.0215$, respectively.

\subsection{Optimization-protocol ablations}

Table~\ref{tab:adaptive} separates feedback selection, rubric adaptation, and checkpoint selection.
Here, best/best denotes held-out selection at both stages, whereas final/final uses the last checkpoint from each stage.

\begin{table}[ht]
\centering
\caption{Judge-T ablations of feedback selection, rubric adaptation, and checkpoint selection.}
\label{tab:adaptive}
\begin{tabular*}{\columnwidth}{@{\extracolsep{\fill}} l l r r @{}}
\toprule
Rubric & Feedback + checkpoint & R1 & R2 \\
\midrule
Adaptive & All + best/best       & 81.91 & 80.29 \\
Fixed    & Bottom + best/best    & 82.15 & 81.98 \\
Adaptive & Bottom + final/final  & 82.43 & 81.10 \\
Adaptive & Bottom + best/best    & \textbf{84.50} & \textbf{84.38} \\
\bottomrule
\end{tabular*}
\end{table}

Under adaptive training and frozen-rubric selection, restricting updates to the lower-scoring half improves Judge-T R1 from 81.91 to 84.50 and R2 from 80.29 to 84.38 relative to using all document diagnoses.
The paired differences are $+2.59$ [$-0.43$, $5.88$] and $+4.09$ [$1.53$, $6.71$], respectively, suggesting a benefit from focusing feedback on lower-scoring documents, particularly for R2.
Holding bottom-half feedback and checkpoint selection fixed, adaptive training criteria raise R1 from 82.15 to 84.50 and R2 from 81.98 to 84.38.
The paired gains are $+2.35$ [$0.13$, $4.85$] and $+2.40$ [$0.45$, $4.40$].
This comparison shows that the automatically constructed base rubric is already usable, while criteria induced by recurring failures provide additional optimization signal.
Within the adaptive bottom-half run, held-out checkpoint selection improves R1 by 2.07 points and R2 by 3.28 points over using the final
checkpoints, suggesting that the learned selection criterion is broadly aligned with the expert-designed evaluation criteria.

Across five update rounds, the adaptive rubric introduced four supplemental criteria, three of which remained active in the final-round rubric: cross-record reference integrity, experimental scope and field semantics, and output integrity. These criteria respectively target resolvable links across records, correct assignment of results to experimental scope and schema fields, and malformed, truncated, or schema-invalid JSON. A fourth criterion, evidence-and-entity consistency, was introduced in the first round and retired after three inactive rounds. This pattern suggests that optimization primarily exposed failures in cross-record consistency, experimental-context assignment, and output validity, while the retirement mechanism prevented inactive criteria from accumulating.
\subsection{Gain and error analysis}

Across the eight rubric dimensions, Joint-TG matches or exceeds Auto-init under both evaluators. Judge-T reports its largest gains in mechanistic fidelity ($+8.00$), causal linking ($+7.50$), and parameter alignment ($+6.50$); Judge-5.5 reports its largest gains in coverage ($+10.00$), causal linking ($+6.00$), and specificity ($+4.00$). Both judges therefore identify improvements in information recovery and in preserving relations among values, conditions, and scientific claims. The smaller gains in usability ($+1.00/0.00$) indicate that Auto-init already produced structurally usable records.

Document-level diagnostics reveal residual errors masked by the averages. In one dehydrogenation–cracking paper, Joint-TG misidentifies a reaction step and misattributes a literature-derived barrier. In a hexadiene-cyclization study, it omits part of a key energy map. These cases indicate remaining challenges in mechanistic grounding and evidence coverage.
\section{Conclusion}
We study how task-specific configurations for scientific extraction agents can be automatically constructed and optimized from weak specifications. Starting from only a short task goal and unannotated reference documents, the framework
constructs a task-specific schema, extraction instructions, and base training rubrics, then refines the schema-construction policy and extraction instructions through failure-focused optimization
while adapting training-time evaluation criteria to recurring failures. 

On a 50-paper catalysis corpus, automatic construction is competitive with manually specified baselines, and both components remain improvable.
Extraction-instruction optimization yields the larger single-stage gain, while schema-construction optimization provides additional improvement when the two stages are optimized sequentially. Ablations further show gains from adaptive training criteria and held-out checkpoint selection. Joint-TG achieves the highest mean across all four judge–rubric settings, and blinded human evaluation also favors Joint-TG over Auto-init and AgentCAT.
Together, these results support treating task construction itself as an optimization target for scientific extraction from weak specifications.
\vfill\pagebreak
\clearpage
\noindent\textbf{Compliance with Ethical Standards.}
This study uses publicly available scientific literature for which no ethical approval was required.

\noindent\textbf{Acknowledgments.}
This work was supported by the Strategic Priority ResearchProgram of the
Chinese Academy of Sciences (Grant No. XDA0490000).
The authors declare no conflicts of interest.
\bibliographystyle{IEEEbib}
\bibliography{refs}
\end{document}